\documentclass[a4paper,fleqn]{cas-dc}

\usepackage[numbers]{natbib}

\def\tsc#1{\csdef{#1}{\textsc{\lowercase{#1}}\xspace}}
\tsc{WGM}
\tsc{QE}
\tsc{EP}
\tsc{PMS}
\tsc{BEC}
\tsc{DE}

\begin{document}
\let\WriteBookmarks\relax
\def\floatpagepagefraction{1}
\def\textpagefraction{.001}

\shorttitle{DL-Based Auto-Segmentation of PTV for TMLI}

\shortauthors{R.C Brioso et~al.}

\title [mode = title]{Deep Learning‐Based Auto‐Segmentation of Planning Target Volume for Total Marrow and Lymph Node Irradiation}
\tnotemark[1,2]

\tnotetext[1]{This work is supported by Italian Ministry of Health (grant number: GR-2019-12370739, project: AuToMI).}

\author[1]{Ricardo Coimbra Brioso}
\ead{ricardo.brioso@polimi.it}

\author[2,3]{Damiano Dei}

\author[2,3]{Nicola Lambri}

\author[1]{Daniele Loiacono}

\author[2,3]{Pietro Mancosu}

\author[2,3]{Marta Scorsetti}

\affiliation[1]{organization={Dipartimento di Elettronica, Informazione e            Bioingegneria, Politecnico di Milano},
    city={Milan},
    country={Italy}}

\affiliation[2]{organization={Department of Biomedical Sciences, Humanitas University},
    city={Milan},
    country={Italy}}

\affiliation[3]{organization={Radiotherapy and Radiosurgery Department, IRCCS Humanitas Research Hospital},
    city={Rozzano, Milan},
    country={Italy}}

\begin{abstract}
In order to optimize the radiotherapy delivery for cancer treatment, especially when dealing with complex treatments such as Total Marrow and Lymph Node Irradiation (TMLI), the accurate contouring of the Planning Target Volume (PTV) is crucial.
Unfortunately, relying on manual contouring for such treatments is time-consuming and prone to errors. 
In this paper, we investigate the application of Deep Learning (DL) to automate the segmentation of the PTV in TMLI treatment, building upon previous work that introduced a solution to this problem based on a 2D U-Net model. 
We extend the previous research (i) by employing the nnU-Net framework to develop both 2D and 3D U-Net models and (ii) by evaluating the trained models on the PTV with the exclusion of bones, which consist mainly of lymp-nodes and represent the most challenging region of the target volume to segment.
Our result show that the introduction of nnU-NET framework led to statistically significant improvement in the segmentation performance. In addition, the analysis on the PTV after the exclusion of bones showed that the models are quite robust also on the most challenging areas of the target volume.
Overall, our study is a significant step forward in the application of DL in a complex radiotherapy treatment such as TMLI, offering a viable and scalable solution to increase the number of patients who can benefit from this treatment.
\end{abstract}



\maketitle

\section{Introduction}
For several types of cancer, radiotherapy is the most effective treatment, and it is used in more than 50\% of cancer patients either as the main treatment or as a conditioning regimen~\cite{Baskar2012CancerAR}.
Radiotherapy uses ionizing radiation to kill cancer cells and shrink tumors.
To avoid toxicity and side effects, the radiation dose must be delivered to the tumor while sparing the surrounding healthy tissues.
Accordingly, radiation oncologists define the Planning Target Volume (PTV) which is the volume that needs to be irradiated to ensure that the tumor receives the prescribed dose.
Usually, the PTV is defined by adding a margin to the Clinical Target Volume (CTV), to take into account the uncertainties in the treatment delivery and the patient setup.
Similarly, radiation oncologists define the organs at risk (OARs), which are the healthy tissues that must be spared from the radiation dose.
The definition of these volumes requires a time-consuming and error-prone contouring process that is performed by radiation oncologists on the Computed Tomography (CT) scan of the patient.

In the last decade, Deep Learning (DL) proved to be very effective and consistent tool for automating the image contouring process in radiotherapy~\cite{huynh2020artificial,cardenas2019advances,shi2022deep}.
However, despite the availability of several academic and commercial models for contouring the most common OARs and targets~\cite{huynh2020artificial,cardenas2019advances,shi2022deep}, the applications of these techniques to very specific and complex 
  targets, as the one involved by the Total Marrow and Lymph Node Irradiation (TMLI)~\cite{mancosu2020}, are still limited.
Automating the contouring of such complex targets is still an open challenge due to the lack of publicly available datasets, the high variability of the target definition guidelines\cite{cancers15051536}, and the large heterogeneity of the components of the target itself.
Decomposing the target into its components and segmenting them separately is not always a viable solution due to the poor reliability of the annotated contours, making the PTV the only target that meets the clinical quality standards.

In this work, we propose a DL-based approach for automating the segmentation of the PTV in the TMLI treatment. 
In a previous work~\cite{Brioso2023}, we tackled this problem by training a 2D U-Net model for the PTV segmentation, achieving promising results.
In this work we extend our previous work in two major ways: (i) we employed the nnU-Net framework~\cite{nnUnet} to train a 2D U-Net model and a 3D U-Net model for the PTV segmentation, 
  and (ii) we assessed the quality of the PTV segmentation on the most challenging components of the target, by excluding the bones from the volume during the evaluation.
Our results showed that both the 2D and the 3D models trained using nnU-Net outperformed the vanilla 2D U-Net model introduced in~\cite{Brioso2023}, achieving the same median Dice Similarity Coefficient (DSC) respectively of 88.6\% with respect to the 
  85.4\% of the vannila U-Net model.
Similar results were observed for the Hausdorff Distance, with 2D and 3D nnU-Net models achieving a median HD95 respectively 
  of 6.6mm and 6.8mm, while the vanilla U-Net model achieved a median HD95 of 11.6 mm.
Even more interestingly, the performance of all the models does not degrade excessively when excluding the bones from the PTV, leading to a decrease of median DSC that ranges from 2.6\% for the 3D nnU-Net model to 3.2\% for the other two models.
Our statistical analysis showed that the differences of performances between the nnU-NET models and the vanilla U-NET model 
  are statistically significant, while we found no statistically significant differences among the two nnU-NET models trained.
Overall, our results suggest that employing Deep Learning models to assist radiation oncologists in the contouring of the PTV for TMLI is a promising direction and might reduce significantly the time required to create the treatment plan and the inter-observer variability.

\section{Related Work}

Early works on auto-contouring emphasized the use of deformable image registration (DIR), probabilistic, and atlas-based methods~\cite{langerak2013multiatlas, fritscher2014automatic, iglesias2015multiatlas, sharp2014vision, brock2017use, qazi2011autosegmentation, rigaud2019deformable}. In recent years, deep learning (DL) has emerged as a highly promising approach for creating auto-contouring tools effectively utilized in RT planning~\cite{wong2021implementation, harrison2022machine}. Numerous studies have successfully applied deep learning to auto-contouring of OARs and target areas in various anatomical regions, such as the head and neck~\cite{Liu2020, Cardenas2020}, thorax~\cite{Yang2020, Feng2019}, abdomen~\cite{Tong2020, Kim2020}, and pelvis~\cite{Ma2022, Kalantar2021, Dong2019}.

Total marrow irradiation (TMI) poses a significant challenge as it requires delineation of OARs and targets across the entire patient's body. Regarding the segmentation of OARs, a few recent studies have demonstrated the feasibility of developing robust solutions for auto-contouring OARs in whole-body CT images. A notable example is the work of Chen et al.~\cite{CHEN2021175}, who developed WBNet, a system comprising multiple segmentation models~\cite{unet, Tang2019, 10.1007/978-3-319-46723-8_49} to segment 50 OARs across whole-body CT images. Zhou~\cite{Zhou2020} compared 2D and 3D convolution neural network (CNN) approaches in segmenting 17 structures in CT images, finding that the 3D approach achieved an average Dice Score (DSC) of 0.79, while the 2D approach reached 0.67. However, this improvement came at the cost of increased training and inference time. The framework used for 3D segmentation was divided into two steps: one that detected the volume of interest (VOI), a 3D bounding box containing the structure, and another that segmented the structure by focusing on the bounding box. Wasserthal et al.~\cite{Wasserthal2022} introduced a new model based on nnU-Net~\cite{nnUnet} to segment 104 anatomical structures (OARs, bones, muscles, and vessels) in whole-body CT images, achieving a DSC higher than 90\% for most structures. Additionally, commercial solutions such as the one developed by RayStation \cite{raystationSegmentation} and by Limbus AI \cite{limbus} are now available for segmenting OARs in CT images.

On the other hand, there are only a few authors that proposed approaches to segment the target of TMI.
Shi et al.~\cite{shi2022} proposed a dual encoder network that combines the Swin Transformer\cite{SwinTransformerLiu2021} and a ResNet~\cite{He2015}, and then uses a decoder to obtain the segmentation of the lymph nodes and bones in the CT image, which is an integral part of PTV. This new architecture achieved slightly better performance than the U-Net, U-Net++, and DeepLabV3+ in the segmentation tasks.
Additionally, Watkins et al.~\cite{Watkins2022} trained a commercial system based on U-Net to segment the PTV for TMI. Their model was trained to segment PTV-Bone, PTV-Lymph Nodes, PTV-ribs, and PTV-skull, achieving DSC of 0.851, 0.830, 0.946, and 0.814, respectively.

As explored in the rest of the article, the lymph nodes are harder to segment than the rest of the structures contained in the PTV, so analyzing articles that have focused on lymph nodes is important as well.

In \cite{Li2021} it was applied a weakly-supervised strategy that used available lymph node annotations that were created according to the RECIST guidelines \cite{Eisenhauer2009}, turned these annotations into a bounding box and a reinforcement learning framework would move and resize the bounding box to improve the semantic segmentation of an U-Net.

To tackle the similarity of the lymph nodes to other structures, \cite{Bouget2023} used 3D variations of U-Net architecture coupled with anatomical priors of structures that could be used to guide the lymph node segmentation. This approach improved the lymph node segmentation significantly compared with the plain U-Net.

With the goal of auto-delineation of the lymph nodes of the head and neck, in \cite{Weissmann2023} an ensemble of a 3D and a 2D nnU-Net was trained using 35 CT volumes. Three clinical experts compared the auto-delineation to the RO's delineation and quantitatively scored both approaches similarly. 

The delineation of head and neck lymph nodes in \cite{Costea2023} used
several ATLAS-based auto-contouring techniques and DL techniques and compared them, the DL models required less corrections by the RO's and reduced significantly their workload by providing usable models.

In the literature is also possible to observe qualitative analysis done with the help of healthcare professionals to analyse the viability of the application of these models. For example, in \cite{Mikalsen2023} it is shown that 93\% of the OARs contoured by DL techniques did not need any correction, while the automatic contours for CTV needed corrections in general.

Overall, research in this topic is promising and achieving reliable performances in OARs for clinical application, but there is still a way to go to improve full-body CTV, lymph node and PTV automatic segmentation to minimize even further the ROs' workload.

\section{Total Marrow and Lymph Node Irradiation}

Total Body Irradiation (TBI) is an RT technique that exposes the patient's entire body to high doses of radiation. This procedure is effective in reducing in treating almost every type of cancer and reduces relapse of the disease but it comes at the cost of damaging healthy tissue, causing infertility and other symptoms. ~\cite{mancosu2020} \cite{Kerbauy2023}. 

The development of Intensity Modulated Radiotherapy (IMRT) in the 1980s \cite{Brahme1982} \cite{Huh2020}, allowed the creation of machines that can control the amount of radiation and its location of application to reduce the unwanted toxicity effects. IMRT is characterized by delivering the radiation beams from fixed angles and the ability to adjust the beam's intensity to the tumor's shape and size \cite{DeLaFuente2013} and it can have different delivery modes: static, step-and-shoot, and, sliding-window. Further advancements in the IMRT technique resulted in Volumetric Modulated Arc Therapy (VMAT), which enables the treatment machine to rotate continuously as it hits the target with radiation. There is no consensus on which technique has better radiation dose delivery, and although VMAT can result in a faster delivery time, both techniques require complex and detailed planning. The choice of which technique to use depends on the facility's resources, technical experience, and other circumstances.

With IRMT and VMAT, Total Marrow Irradiation (TMI) and Total Marrow and Lymph node Irradiation (TMLI)~\cite{mancosu2020} can be performed instead of TBI, and provide more precise radiation delivery, focusing on target volumes while minimizing radiation exposure to healthy tissues~\cite{Wong2006}.
As the name suggests, TMI is a treatment that irradiates the patient's bone marrow, the TMLI irradiates the patient's bone marrow, lymph nodes, and spleen.

\begin{figure}
	\centering
		\includegraphics[scale=0.06]{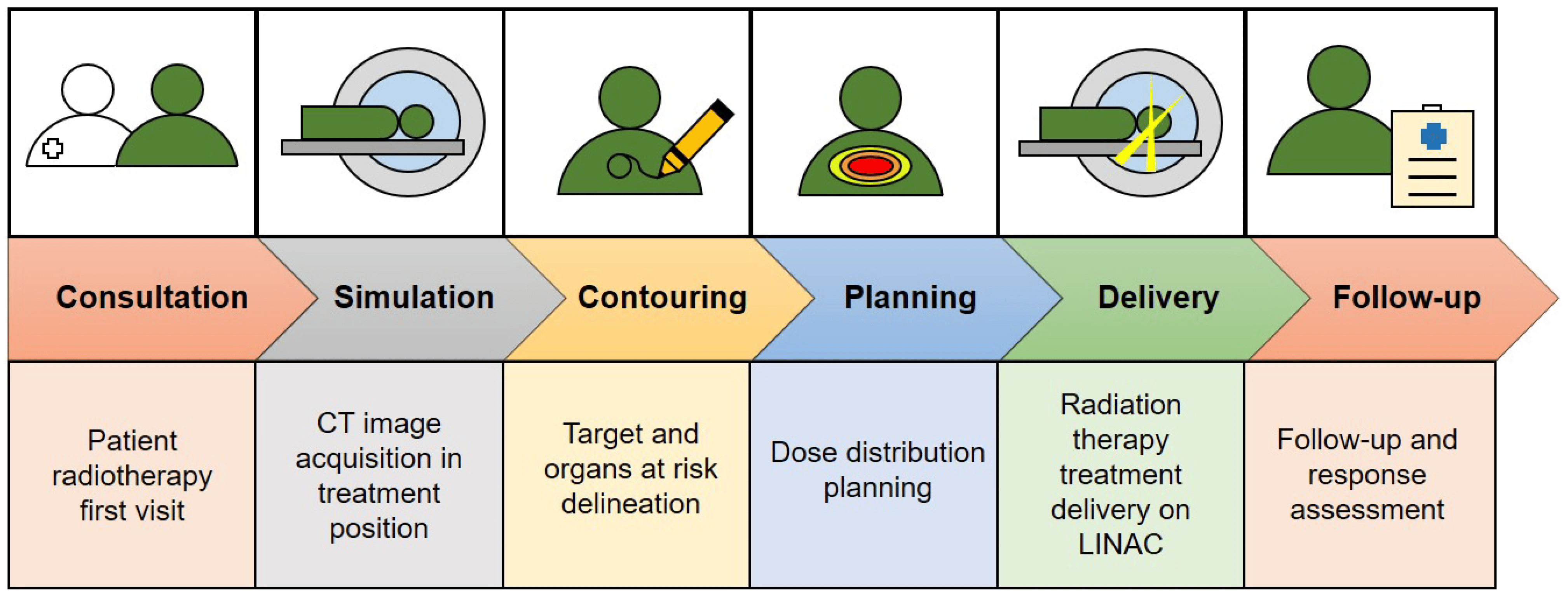}
	\caption{Workflow of the radiotherapy department.}
	\label{fig:temporary_workflow}
\end{figure}

\begin{figure}
	\centering
		\includegraphics[scale=0.5]{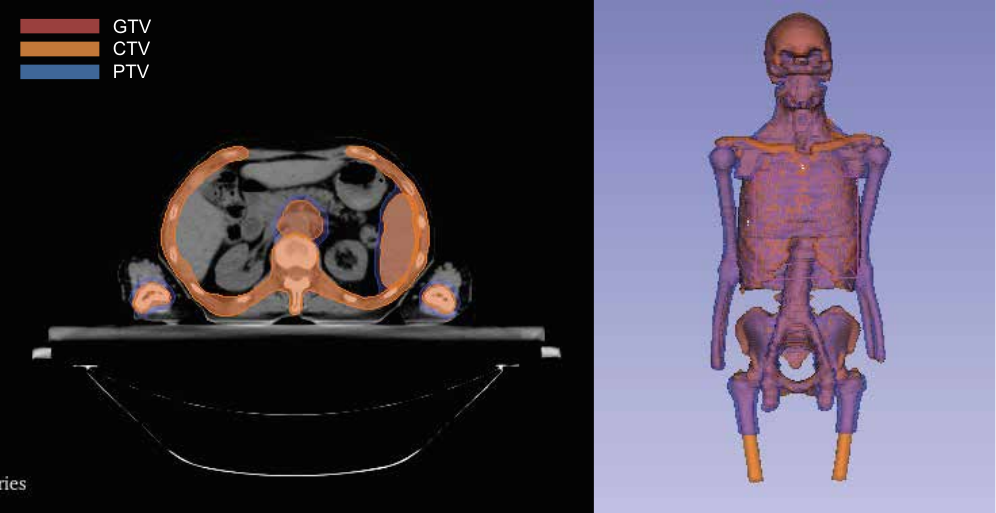}
	\caption{Single CT image of the thoracic area with CTV and PTV labeling done by the RO (left). 3D Representation of the CTV and PTV labeling (right).}
	\label{fig:gtv_ctv_ptv}
\end{figure}

As seen in Figure \ref{fig:temporary_workflow}, the workflow used by the radiotherapy department team starts with patient consultation and image acquisition, secondly, the radiation oncologist (RO) manually segments several structures, including the Organs At Risk (OARs) (lungs, heart, kidneys, etc) and the Gross Target Volume (GTV).
In leukemia treatment, the GTV encompasses bones, marrow, spleen, and lymph nodes, it is delineated according to what can be visualized as cancerous tissue in the computerized tomography (CT) scan. The CTV is the Clinical Target Volume (CTV) obtained from the GTV by adding a margin that considers the potential spreading and uncertainty of the tumor. Finally, the PTV is created by adding another margin to the CTV to account for uncertainties in setup positioning, variations in the patient between the initial acquisition and the treatment, and internal motions of the body~\cite{burnet2004}, as seen in Figure \ref{fig:gtv_ctv_ptv}.

After the RO contours all the necessary structures, the calculation of the dosage of radiation that each structure (healthy and non-healthy) receives is performed and optimized, creating a plan for the treatment. Once the planning is ready, it is time for the delivery of the treatment to the patient.

The contour quality is crucial for the success of IMRT techniques. To achieve this quality and detail there is a considerable amount of time the ROs dedicate to their annotation, creating an opportunity for automated segmentation to assist in decreasing the contouring time of the ROs, as seen in \cite{Ali2023}.

\section{Methods}

\subsection{Data}

The data used for the following experiments is described in our previous work, \cite{Brioso2023}. The patients were diagnosed with hematological malignancies and, identified as candidates for allogeneic transplantation and received nonmyeloablative conditioning TMLI at Humanitas Research Hospital (Rozzano, Milan, Italy).

All TMLI patients were immobilized in the supine position with their arms along their bodies, utilizing an in-house dedicated frame with multiple personalized masks~\cite{MANCOSU2021e98}. A free-breathing, non-contrast CT scan with a 5-mm slice thickness was acquired for each patient using a BigBore CT system (Philips Healthcare, Best, Netherlands).
For these experiments, we opted to exclude 6 of the 100 patients from the dataset to obtain a greater consistency of the data, since these patients did not include the same target due to the specificity of their treatment.
The dataset used has full-body CT scans of 94 patients or 23237 images.

The target structures of the TMLI treatment in this dataset are the union of the bone marrow (CTV\_BM), spleen (CTV\_Spleen), and all lymph node chains (CTV\_LN). The CTV\_BM was considered equivalent to the skeletal bones, with the chest wall added to the ribs to account for breathing motions. To minimize oral cavity toxicity, the mandible was excluded from the CTV\_BM, along with the hands, which have an extremely limited bone marrow presence.
The total planning target volume (PTV\_Tot) was defined as the union of three PTVs, derived from the isotropic expansion of three corresponding CTVs, as follows: (i) PTV\_BM = CTV\_BM + 2 mm (+8 mm for arms and legs) to account for setup margin; (ii) PTV\_Spleen = CTV\_Spleen + 5 mm to account for breathing motions and setup margin; and (iii) PTV\_LN = CTV\_LN + 5 mm to account for target residual motion and setup margin.

\subsection{Architecture and Training Details}

The number of axial slices per patient varies within a range from 164 to 534 and the slices are sized at 512 x 512 pixels with a resolution of 1.2 mm x 1.2 mm.
The pixel values of the slices, which represent tissue radio-intensity, are expressed in Hounsfield Units (HU).

In the first approach, to enhance the contrast of all anatomical structures encompassed within the PTV, a linear look up table (LUT) was applied to each pixel value, ranging from -160 HU to 240 HU, as suggested by the RO. This LUT can help highlight the bone tissue while making the lymph nodes more visible. Each slice was encoded as an 8-bit image, with pixel values ranging from 0 to 255.

The first segmentation model used in this work is a U-Net \cite{unet}.
The network receives as input a $512\times512$ image of a CT slice and outputs a $512\times512$ segmentation mask.
It consists of the repeated application of two convolutions with a kernel of $3\times3$ (unpadded convolutions), each followed by a rectified linear unit (ReLU) and a $2\times2$ max pooling operation for downsampling. The expansive path consists of an upsampling of the feature map followed by a $2\times2$ up-convolution that doubles the number of feature channels and then this is concatenated with the corresponding cropped feature map from the contracting path, and the two $3\times3$ convolutions, each followed by a ReLU.
The final layer is a 1x1 convolution that maps each 64-component feature vector to the desired number of classes. The network has 23 convolutional layers and slightly less than 8 million parameters.

A cross-validation technique was applied and five models were created with different sets of data used for training, validation, and testing.
The sets of all the cross-validation configurations have a similar temporal distribution and the same distribution was used for the remaining experiences.

The other approach consisted of applying the nnU-Net framework \cite{nnUnet} which is an adapting algorithm that analyses the dataset's characteristics, such as its resolution, and pixel spacing, this information is used for the pre-processing, and tuning the training parameters of the nnU-Net. This framework is widely used as a strong baseline for 2D and 3D segmentation tasks.

In our dataset, the average shape of a CT image is $237\times512\times512$, the average spacing of each image is $5mm\times1.171875mm\times1.171875mm$. The nnU-Net's image intensity normalization approach starts by calculating, on the global PTV class, the mean, the standard deviation, the $0.5$, and the $99.5$ percentiles of the intensity values. The percentile intensities are clipped from the image and all of the images are normalized subtracting the global mean and dividing by the standard deviation.

\begin{figure}
	\centering
		\includegraphics[scale=0.2]{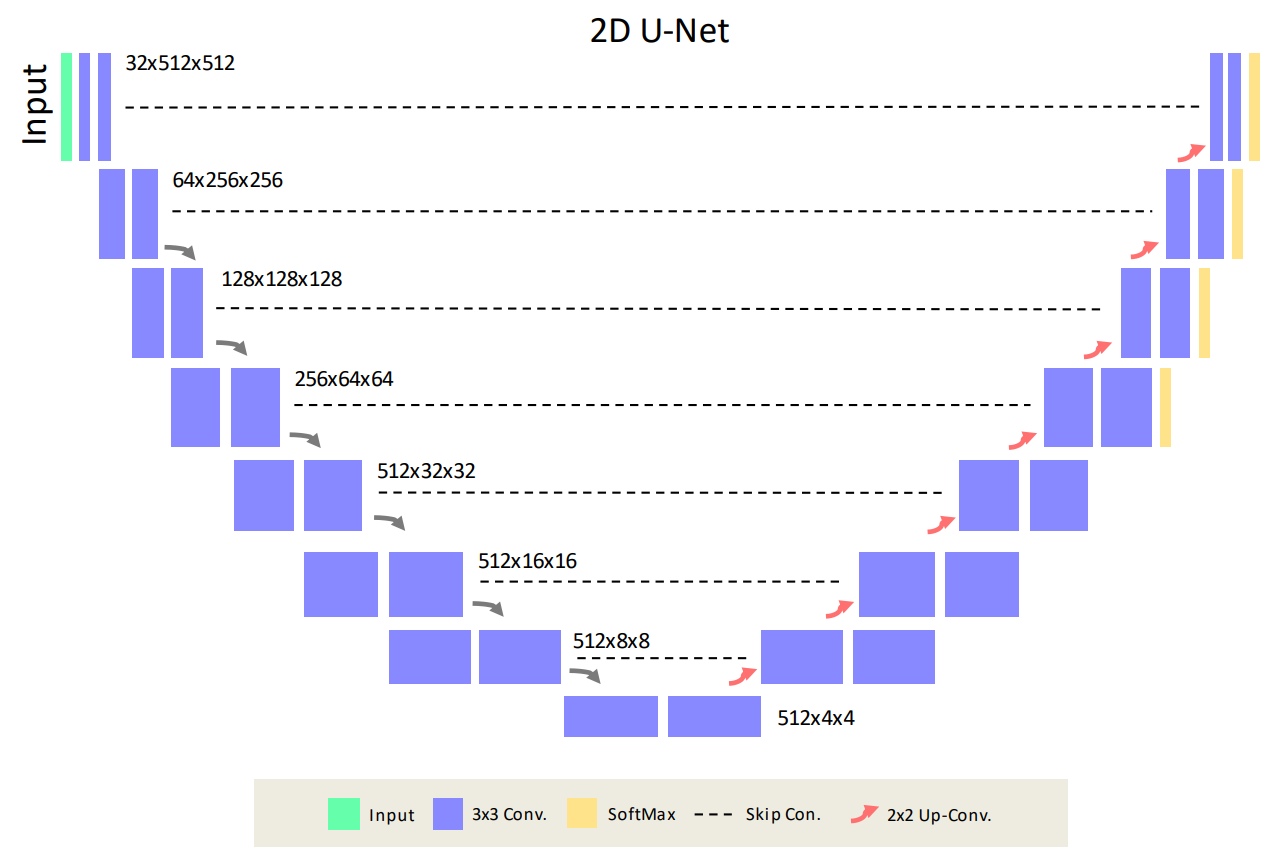}
	\caption{Architecture of the 2D U-Net generated by the nnU-Net framework.}
	\label{fig:nnunet2d_arch}
\end{figure}

In the configuration that the nnU-Net generated, the 2D architecture has 8 encoder blocks and 7 decoder blocks, each block had two $3\times3$ convolutions (as shown in Figure \ref{fig:nnunet2d_arch}, the input shape of each slice was $512\times512$ and the batch size used as 12.

\begin{figure}
	\centering
		\includegraphics[scale=0.25]{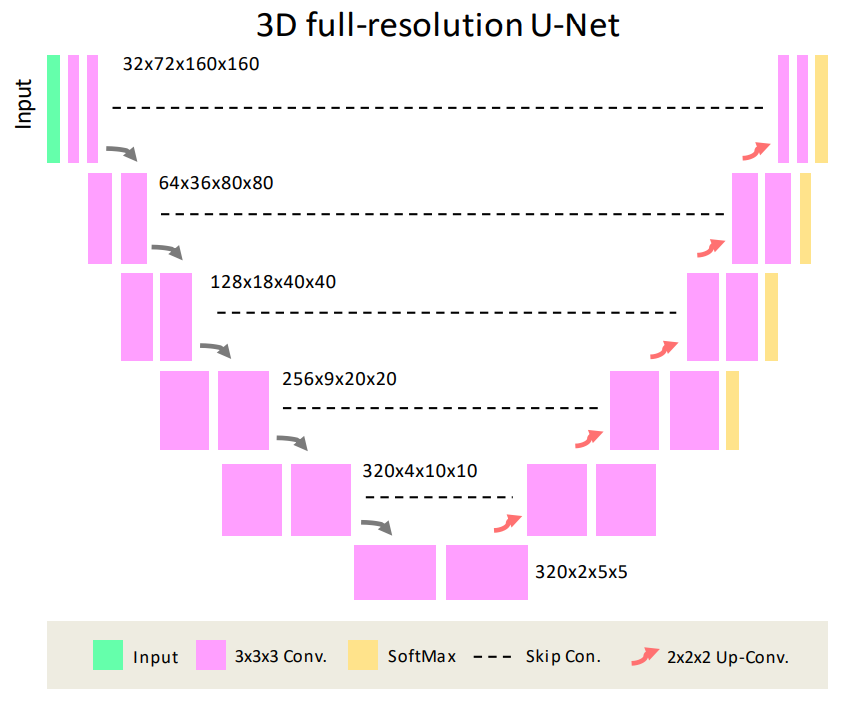}
	\caption{Architecture of the 3D U-Net generated by the nnU-Net framework.}
	\label{fig:nnunet3d_arch}
\end{figure}

For the 3D architecture, as seen in \ref{fig:nnunet3d_arch}, the number of encoder blocks is 6 and the number of decoder blocks is 5, the input is a patch of size $72\times160\times160$ and the batch size is 2.

In \cite{Crespi2022} and \cite{Avesta2023} has been shown that the use of a 3D segmentation model is useful in large datasets or using more complex frameworks, In the first experiment in \cite{Brioso2023}, a normal U-Net was used, as a follow-up experiment in this article, we compared the 2D nnU-Net and the 3D full resolution nnU-Net.

\subsection{Evaluation Metrics}

To quantitatively evaluate the performance of the models, we calculated the Dice Score (DSC) and Hausdorff Distance (HD) as described in Equation \ref{Eq:DSC} and Equation \ref{Eq:HD}, respectively. 
The DSC measures the overlap between the ground truth and the predicted mask, with $X$ representing the set of positive pixels in the ground truth and $Y$ representing the set of positive pixels in the prediction.

\begin{equation}
    DSC = \dfrac{2 |X \cap Y| }{|X| + |Y|}
    \label{Eq:DSC}
\end{equation}

The HD measures the maximum distance of all the nearest distances between the surfaces of the two sets $X$ and $Y$, denoted as $S_X$ and $S_Y$. To mitigate the impact of outliers on HD values, we employed HD95, which excludes the top 5\% highest HD values.

\begin{equation}
    HD=max\left\{\max_{x\in S_X} d(x,S_Y),
    \max_{y\in S_Y} d(y,S_X) \right\},
    \label{Eq:HD}
\end{equation}

To validate the differences in the performance of the models, it was used a paired t-test between the evaluation metrics of the patients' predictions of each model.

\subsection{Assessment of complex structures with the PTV}

The PTV is an agglomerate of structures and margins that are added to these structures. In TMLI, the bones are included in the PTV and are one of the most visible structures in the CT scan. They have high contrast when compared with the rest of the image, and their shape and location are similar from patient to patient. These characteristics of the bones make them more differentiable and easier to segment.

The other structures in the PTV, such as the spleen and the lymph nodes can be confused with muscles or other organs with similar pixel intensity, patterns, and shapes. The lymph nodes can be inflamed or not, changing their size and visibility even within the same patient's CT. Using the evaluation metrics in the entire PTV structure does not show how well the segmentation model performs in the more complex structures.

In this work, we go towards the assessment of the more complex structures in the PTV by using an available open-source model to segment the bones of each patient and subtract them from the PTV.

TotalSegmentator is a segmentation framework implemented as a Python library that provides segmentation models for various structures in CT images, such as bones and several other organs. bone masks of all the patients were created. The bones used in the experiment were: the vertebrae, the ribs, the humeri, the scapulae, the clavicles, the femurs, the hip bone and the sacrum. TotalSegmentator is used for a new evaluation step that can be applied by subtracting the bone segmentation to the PTV groundtruth and the PTV prediction images and re-calculating the quantitative metrics.

\begin{figure}
	\centering
		\includegraphics[scale=0.6]{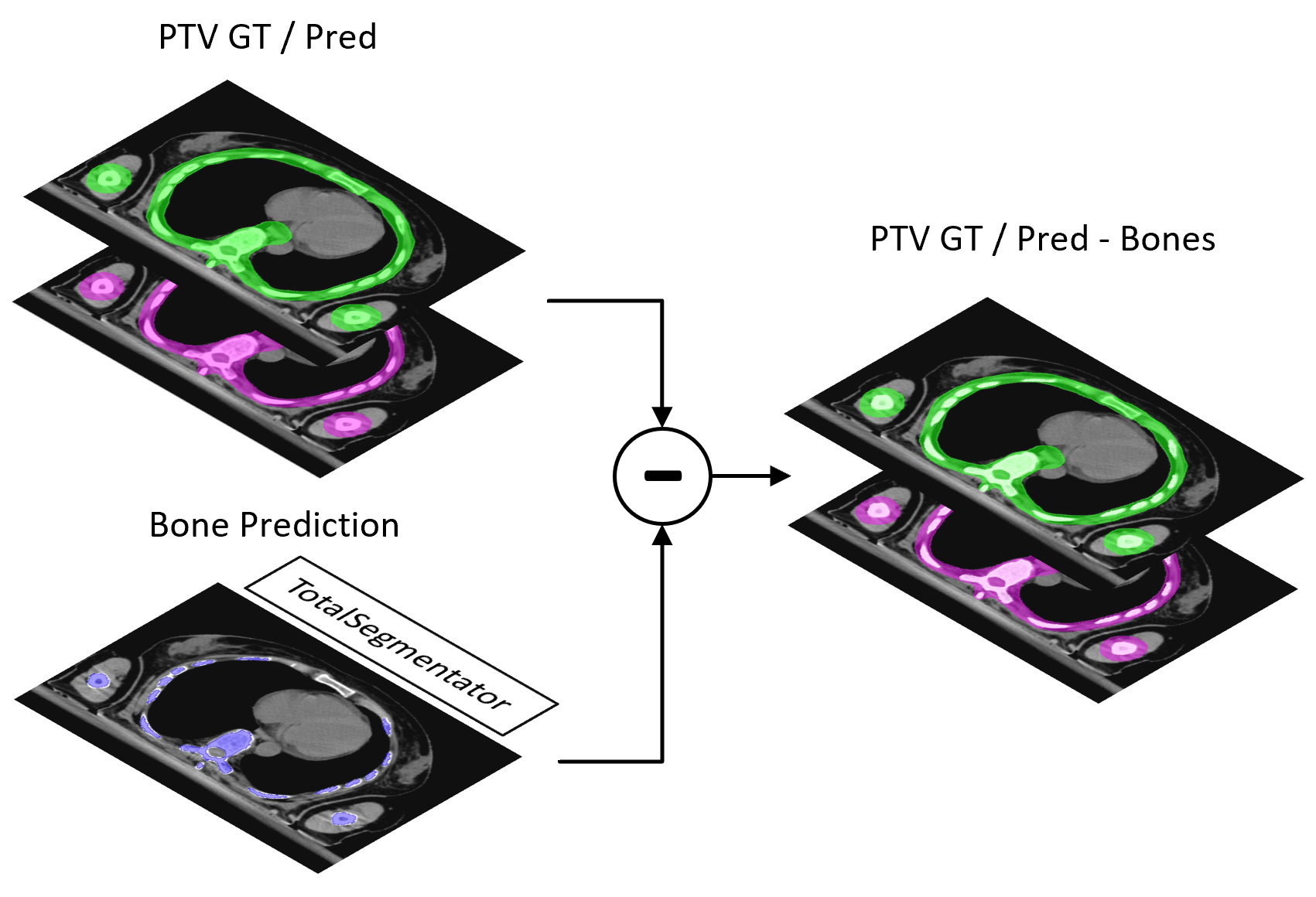}
	\caption{Bone Subtraction Pipeline Description.}
	\label{fig:bone_subtraction_graphic}
\end{figure}

As shown in Figure \ref{fig:bone_subtraction_graphic}, the bone predictions obtained with TotalSegmentator are subtracted to the groundtruth of the PTV (PTV GT) and the prediction of the PTV made by the nnU-Net (PTV Pred). The result of this subtraction is then used to re-evaluate the performance of the PTV segmentation without considering the bones.

\section{Results}

\subsection{Performance Evaluation}
\begin{figure}
	\centering
		\includegraphics[scale=0.60]{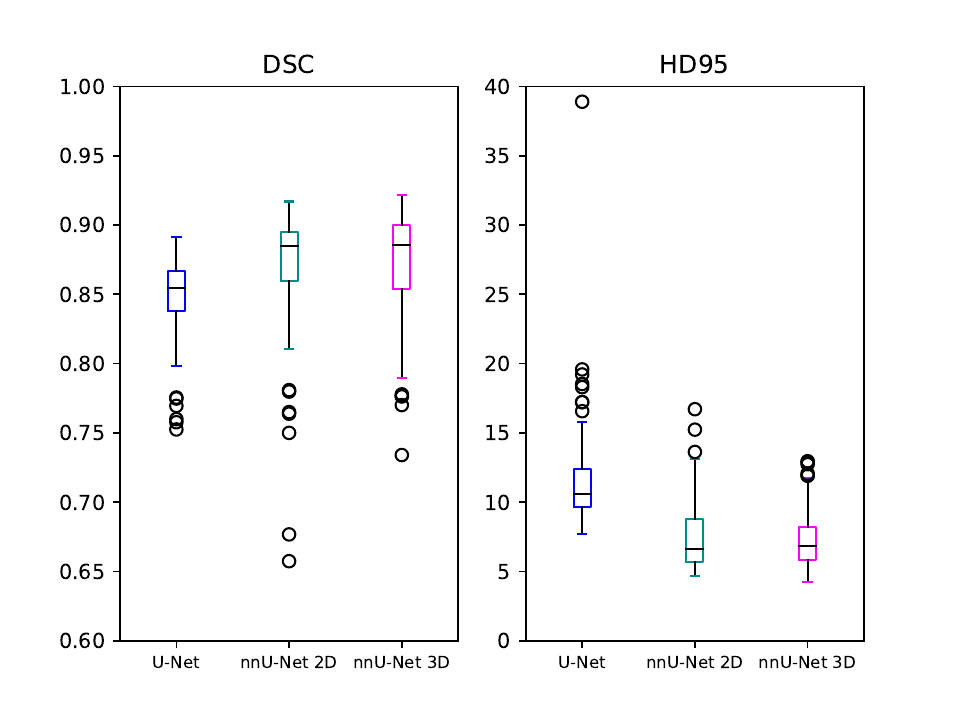}
	\caption{Boxplot of the DSC and HD95 evaluation metrics for the PTV segmentation on the plain U-Net, nnU-Net 2D, and nnU-Net 3D full resolution models.}
	\label{fig:boxplot_unet_nnunet2d_nnunet3d_dsc_hd95}
\end{figure}

We trained and compared three different models for the PTV segmentation: a plain U-Net, a 2D nnU-Net, and a 3D nnU-Net. 
For each model, we trained five different instances using cross-validation, to ensure a proper assessment of the models' performance on all the patients included in the dataset.
Figure \ref{fig:boxplot_unet_nnunet2d_nnunet3d_dsc_hd95} shows the distributions of DSC and HD95 of the three models for the PTV segmentation on the entire dataset.
The plain U-Net model has a median DSC of $85.4\%$, with a range that goes from $75.2\%$ on the worst patient to $89.1\%$ on the best one; the HD95 has a median of $10.6mm$, with a range from $7.7mm$ to $38.9mm$.
The 2D nnU-Net model has a median DSC of $88.6\%$, with a range from $65.7\%$ to $91.7\%$; the HD95 has a median of $6.6mm$, with a range from $4.7mm$ to $16.7mm$.
Finally, the 3D nnU-Net model has a median DSC of $88.6\%$, with a range from $73.4\%$ to $92.2\%$; the HD95 has a median of $6.8mm$, with a range from $4.2mm$ to $13.0mm$.
Accordingly, the results suggest that all the three models achieve a good average performance in the PTV segmentation, with nnU-Net models showing an improvement over the plain U-Net model.
To assess the statistical significance of the differences between the models, we performed a Kruskal-Wallis test, which showed that the differences between the models are statistically significant (p-value $< 0.001$).
Thus we performed the Dunn's post-hoc test with Holm correction to compare the models in pairs, and we found that the differences between the plain U-Net and the 2D nnU-Net are statistically significant (p-value $< 0.001$), as well as the differences between the plain U-Net and the 3D nnU-Net (p-value $< 0.001$). In contrast, the differences between the 2D nnU-Net and the 3D nnU-Net are not statistically significant (p-value $= 0.504$).

\subsection{Performance Evaluation after Bone Subtraction}
\begin{figure}
	\centering
		\includegraphics[scale=0.55]{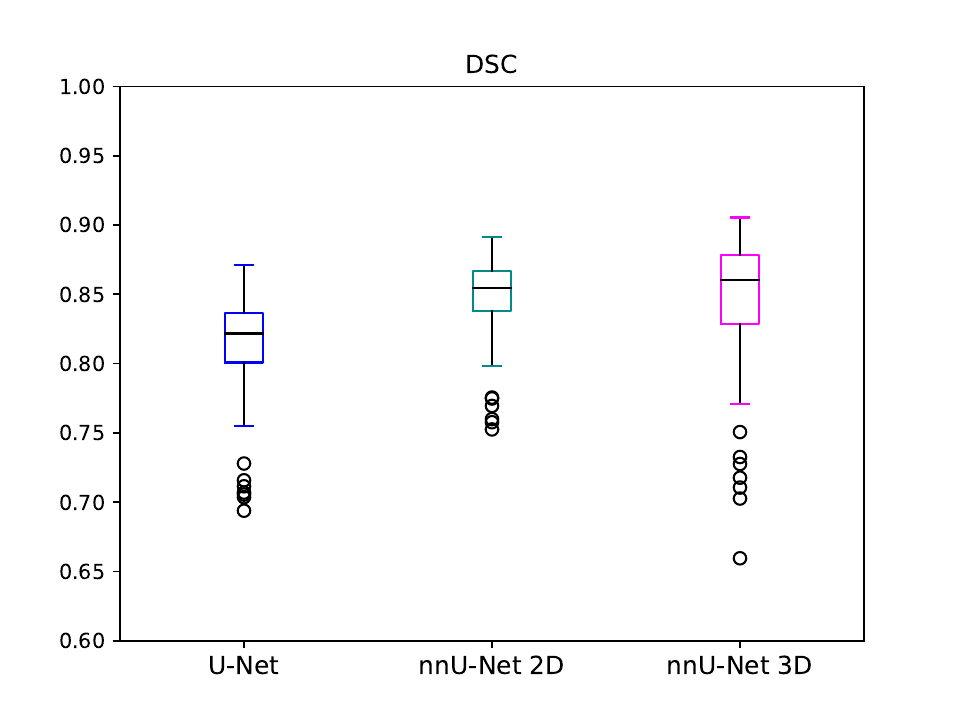}
	\caption{Boxplot of the DSC evaluation metric for the PTV segmentation on the plain U-Net, nnU-Net 2D, and nnU-Net 3D full resolution models after bone subtraction.}
	\label{fig:boxplot_unet_nnunet2d_nnunet2d_nobones_dsc}
\end{figure}

In order to better evaluate the performance of the models on the most challenging components of the PTV, we performed an additional assessment by excluding the bones from the PTV during the evaluation, as described in the previous Section.
Figure \ref{fig:boxplot_unet_nnunet2d_nnunet2d_nobones_dsc} shows the distributions of DSC of the three models for the PTV segmentation after the bone subtraction. 
Please notice that the HD95 metric cannot be used in this assessment, because the subtraction of the bones from the predictions and the groundtruth lead to regions with a large fraction of shared margins, which affect the computation of the HD metrics.
The plain U-Net model has a median DSC of $82.2\%$, with a range that goes from $69.4\%$ on the worst patient to $87.1\%$ on the best one.
The 2D nnU-Net model has a median DSC of $85.4\%$, with a range from $75.2\%$ to $89.1\%$.
Finally, the 3D nnU-Net model has a median DSC of $86.0\%$, with a range from $66.0\%$ to $90.5\%$.
These results suggest that the performance of all the models does not degrade excessively when excluding the bones from the PTV, with the two nnU-Net models showing the best performance in this assessment.
As done before, we performed a Kruskal-Wallis test to assess the statistical significance of the differences between the models, and we found that the differences between the models are statistically significant (p-value $< 0.001$).
Indeed, the Dunn's post-hoc test with Holm correction showed that the differences between the plain U-Net and the 2D nnU-Net are statistically significant (p-value $< 0.001$), as well as the differences between the plain U-Net and the 3D nnU-Net (p-value $< 0.001$). In contrast, the differences between the 2D nnU-Net and the 3D nnU-Net are not statistically significant (p-value $= 0.752$).

\subsection{Qualitative Evaluation}
Finally, we performed a visual evaluation to compare the prediction of the models and to assess the impact of the bone subtraction on the PTV segmentation.

\begin{figure*}
	\centering
		\includegraphics[scale=0.33]{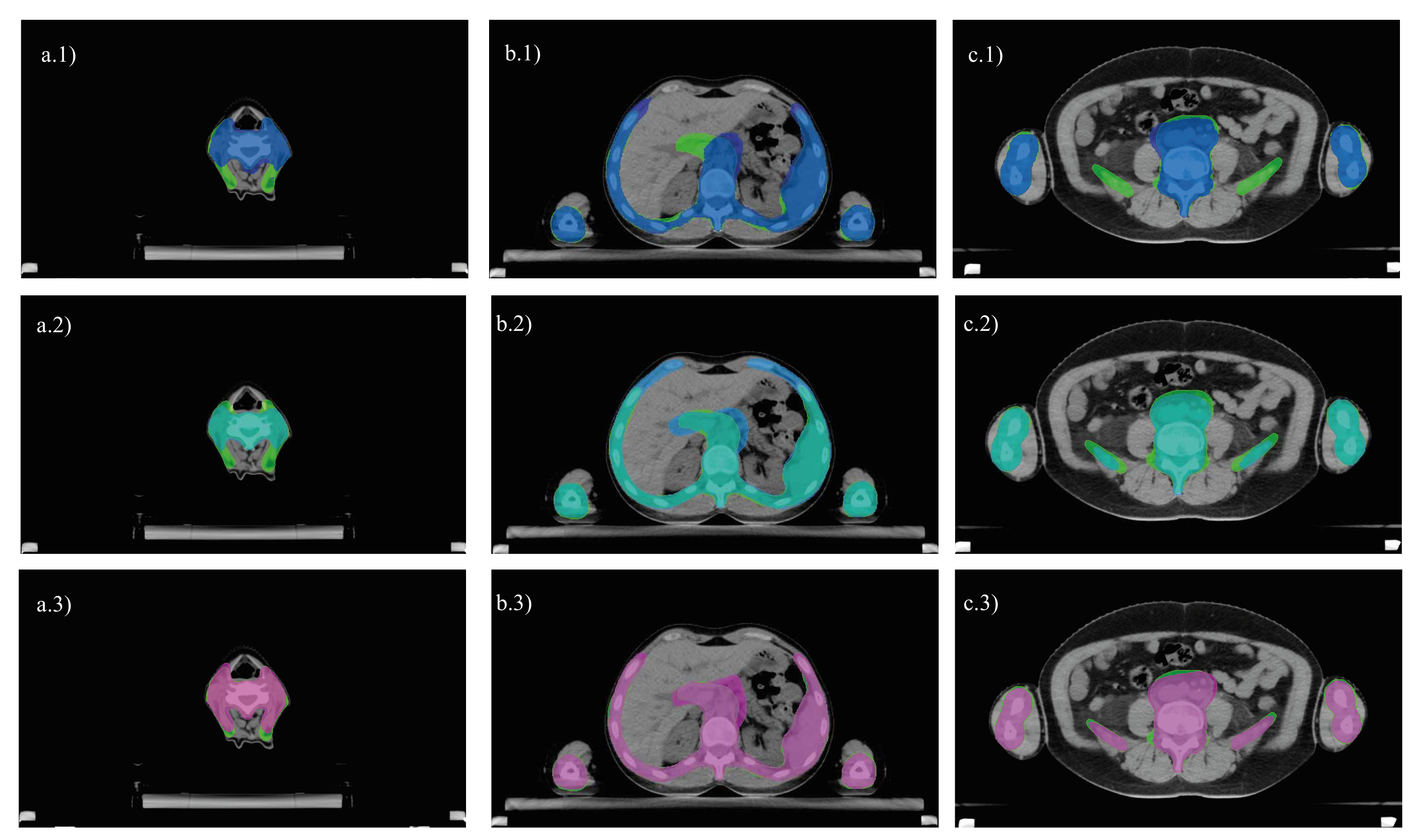}
	\caption{CT slices with overlays of the groundtruth (green) and the predictions of the PTV from the U-Net model (blue), the nnU-Net 2D (light blue) and the nnU-Net 3d full resolution models (magenta).}
	\label{fig:unet_vs_nnunet2d_vs_nnunet3d}
\end{figure*}

In Figure \ref{fig:unet_vs_nnunet2d_vs_nnunet3d}, each column represents a pair of images of one patient and the same slice of the CT volume. The images on the first row contain the overlays of the groundtruth in a green color and the prediction made by the U-Net model in a blue color. The images of the second and third rows, the prediction is instead made by the nnU-Net 2D (light blue), and 3D full resolution (magenta) models, respectively.

In Figure \ref{fig:unet_vs_nnunet2d_vs_nnunet3d}, comparing the first column of images in the head and neck region, we can see that the nnU-Net 3D model covers a bigger area of groundtruth that corresponds to the lymph nodes in that region when compared to the U-Net model and nnU-Net 2D model. The second column of images are slices from the upper body section, the 2D and 3D nnU-Net's prediction covers a greater area of the lymph nodes near the liver compared to the U-Net. Although the 2D nnU-Net's prediction predicts an area of the ribs that is not present in the groundtruth.
In the last column we showcase the region of the hips, the U-Net can not segment properly the hip bone contrary to the 2D and 3D nnU-Net. Between the nnU-Net models, the 3D version displays a more precise segmentation in the last column as well.

\begin{figure}
	\centering
		\includegraphics[scale=0.4]{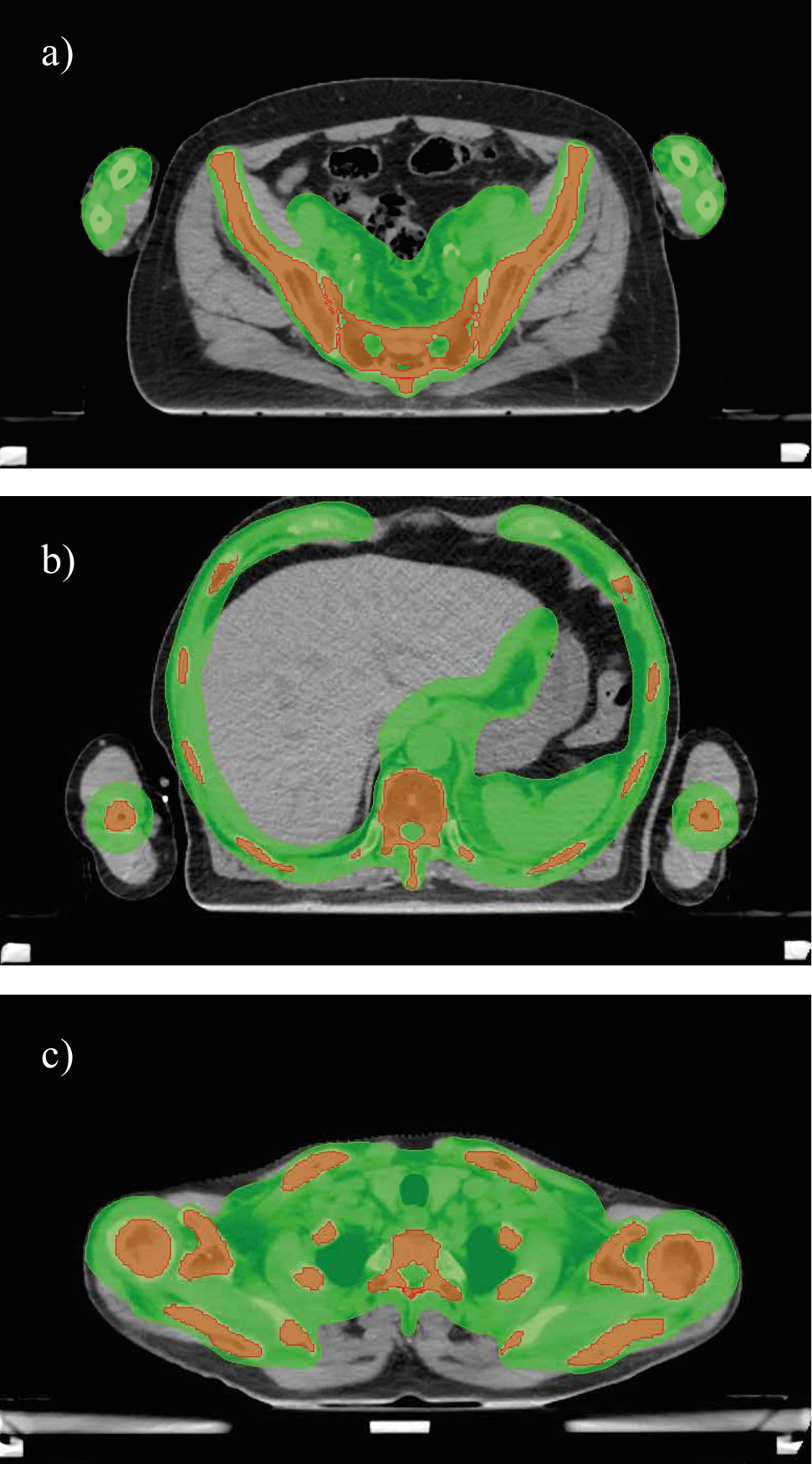}
	\caption{CT slices with overlays of the groundtruth (green) and the predictions of the bones from TotalSegmentator (red).}
	\label{fig:bones_gt}
\end{figure}

The bone predictions obtained with TotalSegmentator can be observed in Figure \ref{fig:bones_gt} and visually evaluating them, they seem accurate, although the version of TotalSegmentator applied in this work did not contain segmentation models for all of the bones. For example, in Figure \ref{fig:bones_gt}.a), we can notice the absence of the segmentation of the radius and the ulna.

\section{Discussion and Conclusion}

In this work, we investigated the application of Deep Learning to segment the PTV in TMLI patients.
In fact, while several commercial and open-source Deep Learning segmentation models are now widely available to assist radiation oncologists with the segmentation of the most common OARs and targets, 
  the PTV of TMLI patients is still a challenge for these models.
In this paper, we extended a previous work\cite{Brioso2023} by comparing the performance of a vanilla 2D U-Net model with both a 2D and a 3D model trained within the nnU-Net framework.
We also investigated the performance of these models on the PTV after the subtraction of the bones from the groundtruth and predictions. This was done to assess the performance of the models on the most 
  challenging region of the PTV, which after the bones exclusion, is mainly composed of the lymph nodes.

Our results suggest that using the nnU-Net framework to train the 3D model significantly improved the performance of the model when compared to the plain 2D U-Net model, leading to the $3.2\%$ increase of 
  the median DSC (from $85.4\%$ to $88.6\%$) and the decrease of the median HD95 from 10.6mm to 6.6mm/6.8mmm (for the 2D and 3D model, respectively).
As expected, the performance of the models decreases when evaluated on the PTV with the exclusion of the bones. 
However, the decrease is not dramatic, with the median DSC decreasing of a $3.2\%$ for the two 2D models and of $2.6\%$ for the 3D model.
This suggests that performance on the whole PTV is not only due to the bones, which are easier to segment but also due to accurate segmentation of the most challenging regions, such as the lymph nodes.

When comparing the 2D and 3D models trained with nnU-Net, we observed that the 3D model has a slightly better consistency across the patients, a slightly more reliable performance on the PTV with bones excluded, and slightly more convincing predictions from a visual inspection.
Nevertheless, the statistical analysis showed that such differences are not significant, suggesting that the 3D model was not able to take advantage of the 3D information to improve significantly the 
  segmentation of the PTV.
This result is not completely unexpected, as it confirms the findings of a previous work~\cite{Crespi2022} that showed how 3D convolutional networks are not able to provide significant improvements over 2D models for the segmentation when the size of the dataset is limited, as is the case of the TMLI dataset considered in this work.

In conclusion, our results are very promising and suggest that Deep Learning models can be effectively used to assist radiation oncologists in complex tasks such as the segmentation of the PTV in TMLI patients.

Future works will focus on developing segmentation models focused only on the lymph nodes, which is the most relevant target from the clinical perspective. 
We also plan to assess the performance of the models through a user study that involves radiation oncologists and medical physicists, to identify how the models can be effectively used in the clinical practice.
Finally, as the contouring of the lymph nodes by the radiation oncologists is greatly enhanced by their previous knowledge of the anatomy of the lymphatic system, we plan to exploit the annotations of the various OARs available in our dataset to enhance the training of more effective segmentation models.





\printcredits

\bibliographystyle{model1-num-names}

\bibliography{paper}

\end{document}